\documentclass[11pt, a4paper]{article}
\usepackage{times, latexsym, url, inconsolata}
\usepackage[T1]{fontenc}
\usepackage{graphicx}
\usepackage{empheq, bm}
\usepackage{booktabs} % professional-quality tables
\usepackage{amsfonts} % blackboard math symbols
\usepackage{nicefrac} % compact symbols for 1/2, etc.
\usepackage{paralist, microtype}
\usepackage{array, multirow}
\usepackage{xcolor, colortbl}
\usepackage{caption}
[font=10pt]
\usepackage{subcaption}
[font=10pt]
\usepackage[all]{nowidow}

\definecolor{BrickRed}{rgb}{0.79,0.25,0.33}
\definecolor{DarkGreen}{rgb}{0.01,0.39,0.01}
\definecolor{DarkBlue}{rgb}{0.01,0.01,0.55}
\definecolor{DarkOrange}{rgb}{0.99,0.55,0.01}
\usepackage[markup=underlined]{changes}
\usepackage{todonotes}
\definechangesauthor[color=BrickRed]{CVN}
\definechangesauthor[color=DarkGreen]{RA}
\definechangesauthor[color=DarkOrange]{RB}
\definechangesauthor[color=DarkBlue]{RC}
\setcommentmarkup{\todo[color={authorcolor!20},size=\scriptsize]{#3: #1}}

\usepackage[acceptedWithA]{tacl2021v1}
\iftaclpubformat % this "if" is set by the choice of options

\else

\fi

\title{A Confidence-based Acquisition Model \\for Self-supervised Active Learning
and Label Correction}

\author{Carel van Niekerk, Christian Geishauser, Michael Heck, Shutong Feng \\ {\bf Hsien-chin Lin, Nurul Lubis, Benjamin Ruppik, Renato Vukovic and Milica Ga\v{s}i\'c}
\\ Heinrich Heine Universität Düsseldorf, Düsseldorf, Germany \\ \texttt{\{cvanniekerk,geishaus,heckmi,fengs,linh,lubis,ruppik,revuk100,gasic\}@hhu.de}
\\}

\newcommand{\NumCategories}{M}
\newcommand{\CategoryIndex}{m}
\newcommand{\EnsembleSize}{E}
\newcommand{\EnsembleIndex}{e}
\newcommand{\NumMetaFeatures}{L}
\newcommand{\MetaFeatureIndex}{l}
\newcommand{\NumValues}{K}
\newcommand{\ValueIndex}{k}
\newcommand{\TotalTimesteps}{T}
\newcommand{\Timestep}{t}

\newcommand{\boldpi}{\bm{\pi}}
\newcommand{\boldpitilde}{\widetilde{\boldpi}}
\DeclareMathOperator{\betazero}{\beta_0}

\newcommand{\boldalpha}{\bm{\alpha}}
\newcommand{\boldalphahat}{\widehat{\boldalpha}}

\DeclareMathOperator*{\argmax}{argmax}

\date{}

\begin{document}
  \maketitle

  % --------------------------------------------------------------------------------------
\begin{abstract}
    % --------------------------------------------------------------------------------------

    Supervised neural approaches are hindered by their dependence on large, meticulously annotated datasets, a requirement that is particularly cumbersome for sequential tasks.
    The quality of annotations tends to deteriorate with the transition from expert-based to crowd-sourced labelling. To address these challenges, we present \textbf{CAMEL} (\textbf{C}onfidence-based \textbf{A}cquisition \textbf{M}odel for \textbf{E}fficient self-supervised active \textbf{L}earning), a pool-based active learning framework tailored to sequential multi-output problems.
    CAMEL possesses two core features: (1) it requires expert annotators to label only a fraction of a chosen sequence, and (2) it facilitates self-supervision for the remainder of the sequence.
    By deploying a label correction mechanism, CAMEL can also be utilised for data cleaning. We evaluate CAMEL on two sequential tasks, with a special emphasis on dialogue belief tracking, a task plagued by the constraints of limited and noisy datasets.
    Our experiments demonstrate that CAMEL significantly outperforms the baselines in terms of efficiency. Furthermore, the data corrections suggested by our method contribute to an overall improvement in the quality of the resulting datasets.\footnote{The code is available under \url{https://gitlab.cs.uni-duesseldorf.de/general/dsml/camell.git}.}
\end{abstract}
  % --------------------------------------------------------------------------------------
\section{Introduction}
\label{section:introduction}
% --------------------------------------------------------------------------------------

% Introduction
Supervised training of deep neural networks requires large amounts of accurately annotated data~\cite{russakovsky2015imagenet,szegedy2017inception,li2020unified}.
A particularly challenging scenario arises when training for sequential multi-output tasks.
In this case, the neural network is required to generate multiple predictions simultaneously, one for each output category, at every time step throughout an input sequence.
Consequently, the labelling effort increases rapidly, becoming impractical as the demand for precise and consistent labelling across each time step and output category intensifies.
Therefore, a heavy dependence on human-generated labels poses significant limitations on the scalability of such systems.

A prominent example of a sequential multi-output label task for which this bottleneck is evident is dialogue belief tracking.
A dialogue belief tracker is one of the core components of a dialogue system, tasked with inferring the goal of the user at every turn~\citep{young2007his}.
% State of the art and its problems
Current state-of-the-art trackers are based on deep neural network models~\citep{lin2021t5dst,vanniekerk2021setsumbt,heck2022trippier}.
These models outperform traditional Bayesian network-based belief trackers~\cite{young2010hidden, thomson2010bayesian}.
However, neural belief trackers are greatly hindered by the lack of adequate training data.
Real-world conversations, even those pertaining to a specific task-oriented domain, are extremely diverse.
They encompass a broad spectrum of user objectives, natural language variations, and the overall dynamic nature of human conversation.
While there are many sources for dialogue data, such as logs of call centres or virtual personal assistants, \emph{labelled} dialogue data is scarce~\citep{vukovic-etal-2022-dialogue} and several orders of magnitude smaller than, say, data for speech recognition~\citep{panayotov2015librispeech} or translation~\citep{bojar2017wmt17}.
Although zero-shot trackers do not require large amounts of labelled data, they typically underperform compared to supervised models that are trained on accurately labelled datasets~\citep{heck2023chatgptdst}.

% Further explanation of the problem
One of the largest available labelled datasets for task-oriented dialogues is MultiWOZ, which is a multi-domain dialogue dataset annotated via crowdsourced annotators.
The challenges in achieving consistent and precise human annotations are apparent in all versions of MultiWOZ~\citep{budzianowski2018multiwoz,eric2019multiwoz21,zang2020multiwoz22,han2020multiwoz23,ye2022multiwoz24}.
Despite manual corrections in the most recent edition, model performance has plateaued, not due to limitations in the models, but as a result of data inconsistencies~\citep{li2020coco,feng-etal-2023-chatter,ruppik-etal-2024-local}.

Addressing the omnipresent issue of unreliable labels, as evident in the MultiWOZ dataset, is a common problem that affects the quality and reliability of supervised learning systems.
In order to mitigate these issues and enhance the robustness of model training, we propose a novel methodology.

% The proposal
In this work, we present CAMEL, a pool-based semi-supervised active learning approach for sequential multi-output tasks.
Given an underlying supervised learning model that can estimate confidence in its predictions, CAMEL substantially reduces the required labelling effort.
CAMEL comprises:
\begin{itemize} 
    \item A selection component that selects a subset of time-steps and output categories to be labelled in input sequences by experts rather than whole sequences, as is normally the case.

    \item A self-supervision component that uses self-generated labels for the remaining time-steps and output categories within selected input sequences.

    \item A label validation component which examines the reliability of the human-provided labels.
\end{itemize}

We first apply CAMEL within an idealised setting for machine translation, a generative language modelling task.
CAMEL achieves impressive results, matching the performance of a model trained on the full dataset while utilising less than \(60\%\) of the expert-provided labels.
Subsequently, we apply CAMEL to the dialogue belief tracking task.
Notably, we achieve \(95\%\) of a tracker's full-training dataset performance using merely \(16\%\) of the expert-provided labels.
Additionally, we propose an adaptation of the meta-post-hoc model approach~\citep{shen2023metauq}, tailored for cost-efficient active learning.
We demonstrate that CAMEL, utilising uncertainty estimates from this cost-effective method, exhibits similar performance compared to using uncertainty estimates from a significantly more computationally expensive ensemble of models.

On top of this framework, we develop a method for automatically detecting and correcting inaccuracies of human labels in datasets.
We illustrate that these corrections boost performance of distinct tracking models, overcoming the limitations imposed by labelling inconsistencies.
Having demonstrated its efficacy in machine translation and dialogue belief tracking, our framework holds potential for broad applicability across various sequential multi-output tasks, such as object tracking, pose detection, and language modelling.
  % --------------------------------------------------------------------------------------
\section{Related Work}
\label{section:related_work}
% --------------------------------------------------------------------------------------

% --------------------------------------------------------------------------------------
\subsection{Active Learning}
\label{subsection:related_work:active_learning}
% --------------------------------------------------------------------------------------

Active learning is a machine learning framework that pinpoints scenarios in data that lack representation and interactively queries a designated annotator for labels~\citep{cohn1996active}.
The framework uses an acquisition function to identify the most beneficial data points for querying.
Such a function estimates how performance can improve following the labelling of data.
Functions of this kind often rely on various factors, such as prediction uncertainty~\citep{houlsby2011bayesian}, data space coverage~\citep{sener2017geometric}, variance reduction~\citep{johansson2007importance}, or topic popularity~\citep{iovine2022popularal}.

Active learning approaches can be categorised into stream-based and pool-based~\citep{settles2009active}.
Stream-based setups are usually employed when data creation and labelling occur simultaneously.
In contrast, pool-based approaches separate these steps, operating under the assumption that an unlabelled data pool is available.

Active learning has been frequently employed in tasks such as image classification~\citep{houlsby2011bayesian,gal2017deep} and machine translation~\citep{vashistha2022activenmt,liu2018activenmt}.
A noteworthy example in machine translation is the work of~\citet{hu2021phrasenmt}, which enhances efficiency by applying active learning to datasets enriched with frequently used phrases.
While this strategy does reduce the overall effort required for labelling, it inherently limits the scope of the annotator's work to phrases only.
As a result, this method may not support the annotation of longer texts, where understanding the context and nuances of full sentences is crucial.

At the same time, active learning is less prevalent in dialogue belief tracking, with~\citet{xie2018cost} being a notable exception.
Their framework involves querying labels for complete sequences (dialogues) and bases selection on a single output category, neglecting any potential correlation between categories.
Furthermore, this approach does not account for annotation quality problems.

One work that addresses the issue of annotation quality within an active learning framework is~\citet{su2018activereward}.
In that work, stream-based active learning is deployed for the purpose of learning whether a dialogue is successful.
The user-provided labels are validated using a label confidence score.
This innovative learning strategy is however not directly applicable to sequential multi-output tasks, as it does not deal with the sequential nature of the problem.

% --------------------------------------------------------------------------------------
\subsection{Semi-Supervised Learning}
\label{subsection:related_work:semi_supervised_learning}
% --------------------------------------------------------------------------------------

Semi-Supervised Learning (SSL) makes use of both labelled and unlabelled data to improve learning efficiency and model performance.
While SSL traditionally encompasses various approaches, including encoder-decoder architectures, alternative methods incorporate self-labelling or self-supervision to enhance model training with minimal human intervention.

In SSL, a ``pre-trained'' model typically undergoes an initial phase of unsupervised learning, leveraging large volumes of unlabelled data to learn representations.
Subsequently, the model is fine-tuned for specific tasks using labelled data.
This fine-tuning process, especially prevalent in state-of-the-art transformer-based models like RoBERTa~\citep{liu2019roberta}, is integral to semi-supervised learning strategies, serving as an illustration of their practical utility~\citep{vanniekerk2021setsumbt,su2022multi,heck2022trippier}.

Moreover, SSL can utilise self-training techniques, such as Pseudo Labelling and Noisy Student Training, where a ``teacher'' model generates pseudo labels for unlabelled data, which are then used to train a ``student'' model.
In this iterative process, the student assumes the teacher role. This semi-supervised training can improve performance without necessitating extra labels.

The Pseudo-Label method proposed by~\citet{lee2013pseudo} is a straightforward and effective SSL technique where the model's confident predictions on unlabelled data are treated as ground truth labels.
This method has been widely adopted due to its simplicity and effectiveness in various domains.

Recent advances in SSL have focused on methods such as FixMatch~\citep{sohn2020fixmatch}, which simplifies the semi-supervised learning pipeline by combining consistency regularisation and pseudo-labelling.
FixMatch leverages weakly augmented data to predict pseudo labels, and strongly augmented data to enforce consistency.

Additionally,~\citet{xie2020noisystudent} propose the Noisy Student method, which extends the teacher-student framework by adding noise to the student model, thereby improving its robustness and performance. Further,~\citet{kumar2020selftraining} explore the concept of gradual domain adaptation through self-training, where a model is iteratively trained on data that gradually shifts from the source to the target domain.
This approach has been shown to effectively handle large distribution shifts by leveraging intermediate domains to improve generalisation.
% }

In summary, the incorporation of self-supervision and iterative training frameworks in SSL has proven to be highly effective, driving advancements in model performance with minimal labelled data. These methods not only enhance the learning process but also reduce the reliance on extensive labelled datasets, making SSL a crucial area of research in modern machine learning.

% --------------------------------------------------------------------------------------
\subsection{Label Validation}
\label{subsection:related_work:label_validation}
% --------------------------------------------------------------------------------------

The process of manually correcting labels is very tedious and expensive.
As a result, many works focus on learning from imperfect labels, using loss functions and/or model architectures adapted for label noise~\citep{reed2014training,xiao2015learning,sukhbaatar2015training}.
Still, these methods have been unable to match the performance of models trained on datasets that include manually corrected labels.
However, the alternative of automated label validation or correction is often overlooked by such works.
It has been shown that learning from automatically corrected labels, e.g. based on confidence scores, performs better than learning from noisy labels alone~\citep{liu2017self, jiao2019self}.
The major drawback of these approaches is that they frequently rely on overconfident predictions of neural network models to correct labels, which can further bias the model.
  % --------------------------------------------------------------------------------------
\section{CAMEL: Confidence-based Acquisition Model for Efficient Self-supervised Active Learning}
\label{section:camell}
% --------------------------------------------------------------------------------------

\begin{figure*}[t!]
    \centering
    \includegraphics[scale=0.75]{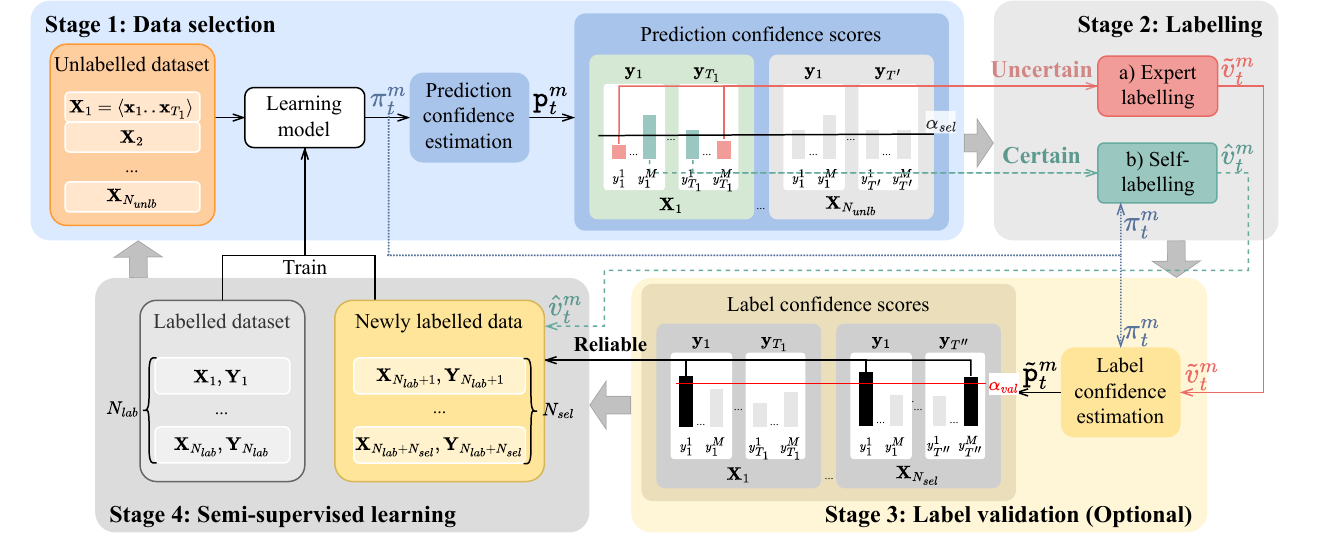}
    \caption{\small CAMEL comprises four stages.
    Stage 1 involves data selection, choosing instances for labelling where the model shows uncertainty (confidence below the $\alpha_{\text{sel}}$ threshold), as indicated by pink arrows.
    In Stage 2, annotators label the selected instances while the model self-labels the remaining ones (dashed green arrows).
    Stage 3 (optional) validates labels using a label confidence estimate, incorporating only labels exceeding the $\alpha_{\text{val}}$ threshold and the self-labelled data into the dataset (black arrows).
    Finally, Stage 4 involves retraining the model for the next cycle.}
    \label{figure:camell:stages}
\end{figure*}

In this section, we introduce our pool-based active learning approach, named \emph{CAMEL}, to address sequential multi-output classification problems.
Let us consider a classification problem with input features $\bm{x}$, and output $\bm{y}$.
According to \citet{read2015multioutput}, such a problem can be cast as a \emph{multi-output} classification problem if the output consists of multiple label categories that need to be predicted simultaneously.
Specifically, for a problem with $\NumCategories$ categories, the output is represented as $\bm{y} = \langle y^{1}, y^{2}, \ldots, y^{\NumCategories} \rangle$, where each $y^{\CategoryIndex}, \CategoryIndex \in [1, \NumCategories]$ can be binary or multivariate.
Furthermore, this problem is characterised as a \emph{sequential} classification problem if the output is dependent on a sequence of prior inputs.
For a sequence with \(\TotalTimesteps\) time-steps, the input-output pairs can be represented as
$\langle ( \bm{x}_{1}, \bm{y}_{1} ), ( \bm{x}_{2}, \bm{y}_{2} ), \ldots, ( \bm{x}_{\TotalTimesteps}, \bm{y}_{\TotalTimesteps} ) \rangle$, 
where $\bm{y}_\Timestep = \langle y_t^{1}, y_t^{2}, \ldots, y_t^{\NumCategories} \rangle$ represents the output labels at time step \(\Timestep \in \left[ 1, \TotalTimesteps \right]\).

In a conventional setting, for an unlabelled data sequence $\bm{X}_i = \langle \bm{x}_1, \ldots, \bm{x}_{T_i} \rangle$, an annotator would typically be required to provide labels, $y_\Timestep^{\CategoryIndex}$, for each label category $\CategoryIndex$ at every time step $t$, which is considerably expensive.

% --------------------------------------------------------------------------------------
\subsection{Requirements}
\label{subsection:camell:requirements}
% --------------------------------------------------------------------------------------

CAMEL, as a confidence-based active learning framework, utilises confidence estimates to determine data points to be queried for labelling.
The framework relies on the model's ability to gauge the certainty of each prediction.
Specifically, for every time-step $t$ in a sequence, for each category $m$ in a multi-output setting, and for each possible value $v \in \mathcal{V}^m$ that $m$ can take, the model calculates the predictive probability, $\pi^m_t(v)=\texttt{p}\left(y^m_t = v\right)$.
These probabilities, collected into a distribution $\bm{\pi}_t^m = [\pi^m_t(v)]_{\forall v \in \mathcal{V}^m}$, form the predictive distribution that CAMEL uses for active learning decisions.

The calibration of these confidence estimates is also critical.
Calibration refers to the alignment between the model's estimated confidence and the empirical likelihood of its predictions~\citep{desai2020calibration}.
Should the model's confidence estimates be poorly calibrated, it may select instances that are not informative, resulting in an inefficient allocation of the annotation budget and potentially suboptimal performance.

% --------------------------------------------------------------------------------------
\subsection{Active Learning Approach}
\label{subsection:camell:active_learning}
% --------------------------------------------------------------------------------------

The approach we propose starts with an initial learning model, which is trained using a small labelled \emph{seed} dataset and iteratively progresses through four stages: data selection, labelling, label validation, and semi-supervised learning.
These iterations continue until either a pre-defined performance threshold is achieved or the dataset is fully labelled.
The schematic representation of this approach is illustrated in Figure~\ref{figure:camell:stages}.

\paragraph{Stage 1: Data selection} In each cycle, we select a subset of $N_{\text{sel}}$ sequences from the unlabelled pool of size $N_{\text{unlb}}$.
Selection is based on the model's prediction confidence, $\texttt{p}_t^m$ (which will be specified in Equation~\ref{equation:camell:confidence_estimation}).
Instances in which the model displays low confidence (confidence below a threshold $\alpha_{\text{sel}}$) are selected.
More precisely, an input sequence is selected if the model shows high uncertainty for at least one time-step $t$ and label category $m$ instance $y_t^m$.
The $\alpha_{\text{sel}}$ threshold is set such that $N_{\text{sel}}$ sequences are selected for labelling.

\paragraph{Stage 2: Labelling} In the input sequences selected in Stage 1, the learning model self-labels the time-steps and categories, \(\hat{v}_t^m = \argmax_{v \in \mathcal{V}^m}(\pi_t^m(v))\), where its confidence is above the threshold $\alpha_{\text{sel}}$.
Concurrently, expert annotators are responsible for labelling the remaining time-steps and categories.
These labels are denoted by $\tilde{v}_t^m$.

\paragraph{Stage 3: Label validation} This is an optional step, and the variant of CAMEL that contains this stage we call Confidence-based Acquisition Model for Efficient Self-supervised Active Learning with Label Validation (CAMELL). We can consider the labels, $\tilde{v}_t^m$, with label confidence, $\tilde{\texttt{p}}_t^m$, below a threshold $\alpha_{\text{val}}$ to be potentially incorrect.
This label confidence is not assigned by the annotators themselves but is computed by the learning model.
To safeguard the model from being trained with these potentially erroneous labels, we purposely exclude them (i.e., these labels are masked in the dataset).
The $\alpha_{\text{val}}$ threshold can be set using a development set.

\paragraph{Stage 4: Semi-supervised learning} At each iteration of the active learning approach, the expert provided labels that passed validation (Stage 3) and the self-determined labels from Stage 2 are added to the labelled pool, resulting in \(N_{\text{lab}} + N_{\text{sel}}\) data sequences.
Based on these, the learning model is retrained.

% --------------------------------------------------------------------------------------
\subsection{Confidence Estimation}
\label{subsection:camell:confidence_estimation}
% --------------------------------------------------------------------------------------

To accurately estimate the prediction confidence required in Stage $1$ as well as the label confidence in Stage $3$, we propose a confidence estimation model for each stage.
These models are designed to encapsulate the learning model's confidence by considering both its \emph{total} and \emph{knowledge-based} uncertainties.
\emph{Total} uncertainty captures all uncertainty in the model's prediction, irrespective of the source.
Conversely, \emph{knowledge} uncertainty in a model originates from its incomplete understanding, which occurs due to a lack of relevant data during training, or the inherent complexity of the problem~\citep{gal2016uncertainty}.

Both the prediction and label confidence estimation models share the same objective: to estimate the probability that the value $v_t^m$ for a specific label category $m$ at time-step $t$ is correct.
To provide the training data for these models, we assume that the labels in the labelled pool are correct, as they have already been validated.
Furthermore, we retrain these models whenever more data is labelled.

Both models share the same general structure:
\begin{empheq}{align}
    \begin{split}
        \bm{h}_t^m & = \texttt{Enc}_\texttt{Intra-Cat}(\bm{z}_t^m) \\
        \bm{h}_t & = \texttt{Enc}_\texttt{Inter-Cat}( [ \bm{z}_t^j ]_{j=1}^M ) \\
        \texttt{p}_t^m & = \texttt{Conf} (\bm{h}_t^m, \bm{h}_t),
    \end{split}
    \label{equation:camell:confidence_estimation}
\end{empheq}
where $\bm{z}_t^m = [\pi_t^m(v_t^m), \mathcal{T}(\bm{\pi}_t^m), \mathcal{K}(\bm{\pi}_t^m)]$ is a set of uncertainty measures for category $m$.
As illustrated in Figure~\ref{figure:camell:uncertainty_features}, these measures consist of the predictive probability specific to $\pi_t^m(v_t^m)$, along with measures of \emph{total} uncertainty, $\mathcal{T}(\bm{\pi}_t^m)$, and \emph{knowledge} uncertainty, $\mathcal{K}(\bm{\pi}_t^m)$, associated with the predictive distribution $\bm{\pi}_t^m$ (See Sections~\ref{subsubsection:experiments:feasibility_study:implementation} and~\ref{subsubsection:experiments:dst_task:learning_model} for concrete implementations).
The \emph{intra-category encoder} is tasked with extracting important category specific features, $\bm{h}_t^m$, from these uncertainties.
Important features across categories, $\bm{h}_t$, are extracted by the \emph{inter-category encoder}\footnote{During label confidence estimation, for categories not selected for labelling, self-labels are used to complete the \emph{inter-category} features.}.
The \emph{inter-category} encoder allows the model to take advantage of any correlations between categories, which was not done by~\citet{xie2018cost}.
Both the \emph{inter-} and \emph{intra-category} encoders consist of linear fully connected layers.
The \emph{confidence estimation} component generates a confidence score, $\texttt{p}_t^m$, reflecting the accuracy of a given category's value.
This component is composed of a linear feature transformation layer, followed by a prediction layer with a Sigmoid activation function.

The design choices for the confidence estimation models were motivated by a desire to capture both \emph{intra-} and \emph{inter-category} uncertainty for reliable confidence estimation.
We observed that excluding \emph{inter-category} features degraded performance, emphasising the importance of incorporating them.

% --------------------------------------------------------------------------------------
\subsubsection{Prediction Confidence Estimation}
\label{subsubsection:camell:confidence_estimation:prediction}
% --------------------------------------------------------------------------------------

The objective of the prediction confidence estimation model is to assess whether the value predicted by the learning model, \(\hat{v}_t^m\), is the ``true'' value, based on the prediction confidence score $\texttt{p}_t^m$.
This model, also known as the confidence-based acquisition model, is used as the selection criterion in Stage $1$.

% --------------------------------------------------------------------------------------
\subsubsection{Label Confidence Estimation}
\label{subsubsection:camell:confidence_estimation:label}
% --------------------------------------------------------------------------------------

The objective of the label confidence estimation model is to determine whether an annotator's label, $\tilde{v}_t^m$, is the ``true'' value, with this decision being based on the label confidence score $\tilde{\texttt{p}}_t^m$.
In~\citet{su2018activereward} the confidence score of the learning model is directly used for both purposes.
We believe this is a suboptimal strategy, because the model has not been exposed to instances of ``incorrect'' labels.
To address this, we generate a \emph{noisy} dataset featuring ``incorrect'' labels for training purposes.

\begin{figure}[t!]
    \centering

    \begin{subfigure}[b]{0.38\columnwidth}
        \centering

        \includegraphics[scale=0.44]{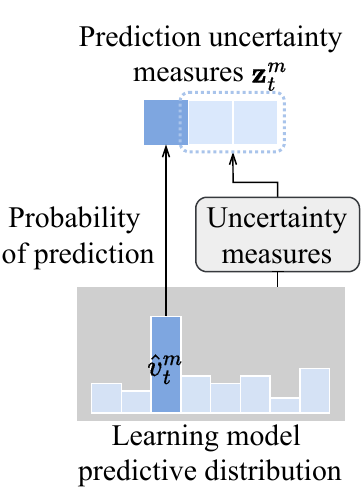}

        \subcaption{\small Prediction}
        \label{subfigure:camell:uncertainty_features:prediction}
    \end{subfigure}
    \begin{subfigure}[b]{0.56\columnwidth}
        \centering

        \includegraphics[scale=0.44]{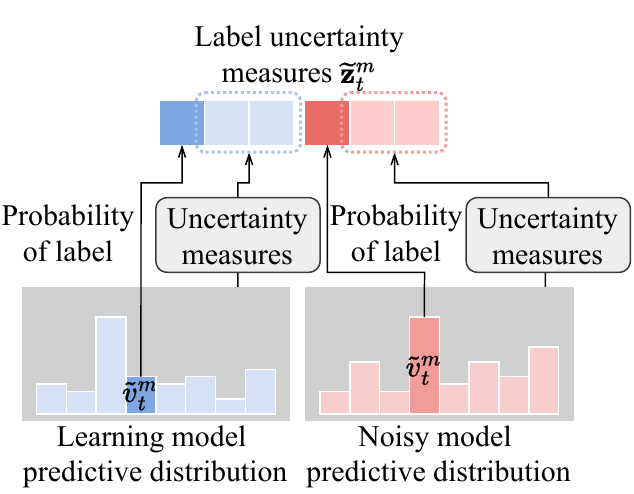}

        \subcaption{\small Label}
        \label{subfigure:camell:uncertainty_features:label}
    \end{subfigure}

    \caption{\small Category-specific uncertainty measures: (a) displays prediction uncertainty, including prediction probability and total and knowledge uncertainty; (b) depicts label uncertainty, including label probability and total and knowledge uncertainty from both learning and noisy models.}

    \label{figure:camell:uncertainty_features}
\end{figure}

Further, we extend $\tilde{\bm{z}}_t^m$ to include uncertainty measures drawn from both a \emph{noisy} model, trained on the corresponding \emph{noisy} dataset, and the original learning model (as depicted in Figure~\ref{subfigure:camell:uncertainty_features:label}).
Given that the \emph{noisy} model is conditioned to accept the ``incorrect'' labels as correct, the discrepancy in uncertainty between the \emph{noisy} model and the learning model enhances the label confidence estimator's ability to identify potentially incorrect labels.

\paragraph{Noisy dataset} The creation of a noisy dataset can be approached in two ways.
One method is to randomly replace a portion of labels.
However, this approach may not yield a realistic noisy dataset, considering human errors are rarely random.
A second approach, particularly when the learning model is an ensemble, as is often the case for uncertainty-endowed deep learning models~\cite{gal2016dropout,ashukha2020pitfalls}, is to leverage individual ensemble members to supply noisy labels (see Section~\ref{subsubsection:experiments:dst_task:learning_model} for details related to an ensemble free approach).
This method may be more effective, given the individual members' typical lower accuracy compared to the ensemble as a whole.

In our proposed approach, we initially select $\alpha_{\text{noise}}$ percent of the sequences from the training data at random.
For each category $m$, we choose a random ensemble member to generate noisy labels.
This ensemble member creates labels at each time step $t$ by sampling from its predictive probability distribution.
To avoid generating labels from the \emph{clean} dataset, the probabilities of these are set to zero prior to sampling.
The noisy dataset is regenerated after each update of the learning model using the updated ensemble members, enhancing diversity of noisy labels.

%--------------------------------------------------------------------------------------
\subsection{Label Correction}
\label{subsection:camell:label_correction}
%--------------------------------------------------------------------------------------

We propose a label correction method that utilises the model that solves the task at hand, referred to as the learning model, the label confidence estimation model (Section~\ref{subsubsection:camell:confidence_estimation:label}) and the prediction confidence estimation model (Section~\ref{subsubsection:camell:confidence_estimation:prediction}).
In order to correct a noisy dataset, this method involves three steps: (1) detecting potentially erroneous labels, (2) determining which of these labels can be accurately corrected by the learning model, and (3) substituting the incorrect labels with the learning model's predictions.
Detecting potentially erroneous labels requires utilising the label confidence estimation model and setting the hyperparameter $\alpha_{val}$, such that all labels $\tilde{v}_t^m$ with confidence below this threshold are considered potentially incorrect.
Then the prediction confidence model is utilised to estimate the learning model's confidence of detected erroneous labels $\tilde{v}_t^m$.
If this confidence is greater than the one assigned by the label confidence estimation model, the labels are substituted with the learning model's predictions.
% --------------------------------------------------------------------------------------
\subsection{Efficient Confidence Estimation with Post-Hoc Uncertainty Learning}
\label{subsection:camell:efficient_confidence_estimation}
% --------------------------------------------------------------------------------------

To obtain reliable estimates of the knowledge and total uncertainties required in Section~\ref{subsection:camell:confidence_estimation}, an ensemble-based approach is typically employed, however, this method is computationally expensive~\citep{gal2016uncertainty}.
This challenge is amplified in active learning scenarios, where the model is frequently updated.
\citet{shen2023metauq} propose an uncertainty estimation technique in which uncertainties are generated by a post-hoc Dirichlet meta-model, offering greater computational efficiency than an ensemble of models.
This method enables the model to distinguish between knowledge and data uncertainty, without needing several instances of the learning model.
The post-hoc Dirichlet meta-model, involves a two-stage training process.
In the initial stage, a model with the same architecture as the learning model is trained to create a base model.
In the second stage, meta-features are employed to estimate the uncertainties of the base model.
These meta-features, derived from various intermediate layers of the base model, capture distinct levels of feature representation, from low- to high-level representations.
Utilising the diversity in these representations allows for more nuanced uncertainty quantification \citep{shen2023metauq}.
To capture the uncertainty of the base model, we utilise a meta-model.
This meta-model takes as input the intermediate features from the base model and outputs the parameters of a Dirichlet distribution.
This Dirichlet distribution over the probability simplex, in turn, describes the uncertainty present in the prediction.

More rigorously, given a base neural network model that solves the task at hand, the set of \(\NumMetaFeatures\)  features \(\bm{F}=\left\{ \bm{f}_1, \bm{f}_2, \ldots, \bm{f}_\NumMetaFeatures \right\}\) is extracted from different layers of this model for a given input, where \(\NumMetaFeatures\) refers to the number of layers of the base model.
These intermediate features can include embeddings from various layers within a neural network, such as the transformer layers in a transformer model.
Meta-features are computed via small meta-feature extraction layers, \(g_\MetaFeatureIndex\).

In our case, these are fully-connected layers with a ReLU activation function that map the intermediate features to meta-features of dimension $d_\texttt{meta}$, \(
\bm{m}_\MetaFeatureIndex = g_\MetaFeatureIndex ( \bm{f}_\MetaFeatureIndex ) \) for \( \MetaFeatureIndex = 1, \ldots, \NumMetaFeatures \). These meta-features are then combined and mapped to the required prediction dimension through another fully-connected layer with ReLU activation.

% --------------------------------------------------------------------------------------
\subsubsection{Learning Objective}
\label{subsubsection:camell:efficient_confidence_estimation:learning_objective}
% --------------------------------------------------------------------------------------

The post-hoc meta-model is trained using the Bayesian matching loss \citep{joo2020bayesianmatching} with the same training dataset as the base model.

The loss for the meta-model, denoted as $\mathcal{L}_\texttt{meta}$, is defined as:
\begin{empheq}{align*}
    \begin{split}
        & \mathcal{L}_\texttt{meta} 
        \left( 
            \bm{\theta}^{(\texttt{meta})}; \mathcal{D} 
        \right) 
        \\
        & = \mathbb{E}_{
            \texttt{p} \left( \bm{x}, y \mid \mathcal{D} \right)
        }
        \left[ 
            \mathbb{E}_{
                \texttt{p} 
                \left( 
                    \boldpi \mid \bm{x}, \bm{\theta}^{(\texttt{meta})} 
                \right)
            } 
            \left[ 
                -\log \texttt{p} \left( y \mid \boldpi \right) 
            \right] 
        \right] 
        \\
        & + \lambda 
        \mathbb{E}_{
            \texttt{p} \left( \bm{x}, y \mid \mathcal{D} \right)
        }
        \left[ 
            D_\texttt{KL} 
                \left[ 
                    \texttt{p} \left( \boldpi | \bm{x}, \bm{\theta}^{(\texttt{meta})} \right) 
                    \Big\Vert 
                    \texttt{p} \left( \boldpi | \bm{\beta} \right) 
                \right] 
        \right].
    \end{split}
\end{empheq}
In this expression, the first term represents the expected negative log-likelihood.
The second term, involving the Kullback-Leibler (KL) divergence, quantifies the deviation of the model's predictive distribution from a Dirichlet prior.
This prior, $\texttt{p} \left( \boldpi \mid \bm{\beta} \right)$, represents our belief about the uncertainty before observing the data.

We can show that an optimal state for this model is reached when the output, $\boldalphahat$, equals the sum of the prior concentration parameters and the scaled one-hot encoded label, 
$\boldalphahat = \bm{\beta} + \frac{1}{\lambda} \bm{y}$.
This mechanism enables the model to adjust its uncertainty by integrating both prior knowledge and the evidence gathered from observed data.
However, the reliance on constant prior concentration parameters, $\bm{\beta}$, introduces a limitation.
Specifically, it encourages the model to generate similar uncertainty estimates across all inputs, irrespective of their complexity.
This, however, leads to a model that is under-confident for inputs it can correctly predict and over-confident for inputs it cannot.
To address this problem, we introduce a distillation approach called \emph{Dynamic Priors} within the Bayesian matching loss framework.
Dynamic Priors adapt at each active learning step by leveraging previous model versions, thereby mitigating the constant prior problem.

% --------------------------------------------------------------------------------------
\subsubsection{Dynamic Priors}
\label{subsubsection:camell:efficient_confidence_estimation:dynamic_priors}
% --------------------------------------------------------------------------------------

Dynamic priors leverage the active learning setting in which we operate.
This setting allows the model to access previous versions of the learning model, which can then be used as priors.
The underlying hypothesis is that replacing the constant prior, as described in Section~\ref{subsubsection:camell:efficient_confidence_estimation:learning_objective}, with a dynamic prior -- one that evolves at each active learning step -- addresses the homogeneity issue discussed above.

More concretely, the prior is predicted from the Dirichlet distributions from the previous model version.
If no previous version is available, such as at the beginning of the active learning process, a small ensemble of models, $ \left\{ \bm{\theta}^{(1)}, \bm{\theta}^{(2)}, \ldots, \bm{\theta}^{(\EnsembleSize)} \right\} $, trained on a small seed-set from the active learning initialisation phase, is used to obtain the initial prior.
It is important to emphasise that only the initial prior is obtained using a small ensemble. In all subsequent updates to the model, the predicted Dirichlet distribution from a single model instance is used as the prior.
By parameterising the Dirichlet prior, $\texttt{p} \left( \boldpi \mid \bm{\beta} \right)$, with the previous model's outputs, our approach dynamically adjusts the prior concentration parameters.
This adjustment not only mitigates the issue of constant priors but also increases the model's ability to produce more accurate uncertainty estimates.

In order to represent the knowledge of the ensemble using a Dirichlet distribution, the ensemble's aggregate predictive distribution,
${\boldpitilde} ( \bm{x} ) 
= 
\frac{1}{\EnsembleSize} 
\sum_{\EnsembleIndex=1}^\EnsembleSize \boldpi^{(\EnsembleIndex)} ( \bm{x} )$, 
and the individual distributions, 
$\boldpi^{(\EnsembleIndex)} ( \bm{x} )
=
\texttt{p} \left( y \mid \bm{x}, \bm{\theta}^{(\EnsembleIndex)} \right)$,
are utilised to compute the prior concentration parameters, $\bm{\beta} ( \bm{x} )$.
Specifically, following \citet{ryabinin2021end2}, we use Stirling's approximation. $\bm{\beta} ( \bm{x} )$ is defined as
$\betazero ( \bm{x} ) \cdot \boldpitilde ( \bm{x} )$, 
where $\beta_0 ( \bm{x} )$ is defined as:
\begin{empheq}{align*}
    \begin{split}
        \beta_0 ( \bm{x} ) & =
        \frac{\NumValues-1}{
            2\sum_{\ValueIndex=1}^\NumValues 
            \widetilde{\pi}_\ValueIndex ( \bm{x} ) \cdot d_k ( \bm{x} )
        } 
        \text{, with}
        \\
        d_k ( \bm{x} ) & =
        \log \widetilde{\pi}_\ValueIndex ( \bm{x} ) 
        -
        \frac{1}{\EnsembleSize} \sum\nolimits_{\EnsembleIndex=1}^\EnsembleSize
        \log \pi_\ValueIndex^{(\EnsembleIndex)} ( \bm{x} ).
    \end{split}
\end{empheq}
In the context of active learning, this approach allows our meta-model to incorporate new labels from annotators while retaining the rich uncertainty estimates derived from the ensemble.
As the learning progresses, the priors are continually updated with predictions from the latest model, ensuring that the uncertainty estimates remain current.

Given the Dirichlet distribution $\texttt{Dir} \left( \boldalpha \right)$ produced by the post-hoc meta-model, the total and knowledge uncertainties can be approximated as follows:
\begin{empheq}{align*}
    \begin{split}
        \mathcal{T} \left( \boldpi \right) = \mathcal{H} \left[ \frac{\boldalpha}{\alpha_0} \right],
    \end{split}
\end{empheq}
\begin{empheq}{align*}
    \begin{split}
        \mathcal{K} \left( \boldpi \right) = \mathcal{T} \left( \boldpi \right) + \sum\nolimits_{k=1}^K \frac{\alpha_k}{\alpha_0} \left[ \psi \left( \alpha_k^* \right) - \psi \left( \alpha_0^* \right) \right].
    \end{split}
\end{empheq}
Here $\boldpi \sim \texttt{Dir} \left( \boldalpha \right)$, $\psi(\cdot)$ represents the digamma function, $\alpha_i^* = \alpha_i + 1$, and $\mathcal{H} [\cdot]$ denotes the entropy of a distribution.
Further, to generate noisy data, predictive distributions are sampled from the Dirichlet distribution to simulate different ensemble members.

Finally, it is important to emphasise the computational efficiency of this approach.
During training, only the parameters of the meta-model, which typically constitute less than $5\%$ of the base model's size, are updated. Additionally, in the inference phase, the meta-model incurs an additional computational cost of approximately $15-20\%$ of the total inference cost, resulting in an overall computationally efficient approach to uncertainty estimation.
  % --------------------------------------------------------------------------------------
\section{Experiments}
\label{section:experiments}
% --------------------------------------------------------------------------------------

% --------------------------------------------------------------------------------------
\subsection{Baselines}
\label{subsection:experiments:baselines}
% --------------------------------------------------------------------------------------

\paragraph{Random selection} randomly selects sequences to be annotated.
Random selection is often used as a baseline for active learning approaches, as it allows us to observe the impact of purely adding more labelled data to our labelled pool without strategically selecting sequences to be labelled.
Its advantage is that it maintains the full data distribution with every selection, thus not creating a bias~\citep{dasgupta2008hierarchical}.

\paragraph{Bayesian Active Learning by Disagreement (BALD)} is an uncertainty-based active learning method which employs knowledge uncertainty as the primary metric for selection~\citep{houlsby2011bayesian}.
This technique has established itself as a strong baseline in various applications.
For instance, in image classification tasks~\citep{gal2017deep} and named entity recognition~\citep{shen2017neral}, BALD has shown notable performance.
Its performance is further enhanced when used in conjunction with ensemble models~\citep{beluch2018ensal}.
Given its widespread adoption and proven efficacy, we see BALD as an ideal baseline.

In our study, we examined two criteria for making the selection decision: one based on the cumulative uncertainty across all time-steps and label categories, and another based on the average uncertainty across categories and time.
Upon evaluation, we observed that the latter criterion yielded superior results, and therefore, adopted it as our baseline, which we refer to as \textbf{BALD}.

We further present an enhanced version of BALD which consists of stages $1$, $2$, and $4$ of our approach as outlined in Section~\ref{subsection:camell:active_learning}, utilising knowledge uncertainty as the \emph{prediction confidence estimate}.
We call this BALD with self-supervision, \textbf{BALD+SS}\footnote{Note that we are not able to combine BALD with label validation as knowledge uncertainty does not provide candidate level confidence scores.}

% --------------------------------------------------------------------------------------
\subsection{Variants of CAMEL}
\label{subsection:experiments:camell}
% --------------------------------------------------------------------------------------

We introduce the following variants to understand the individual and collective contributions of our proposed framework's components.

\paragraph{CAML} \textbf{C}onfidence-based \textbf{A}cquisition \textbf{M}odel for active \textbf{L}earning, represents the foundational layer of our framework, incorporating stages 1, 2a and 4 described in Section~\ref{subsection:camell:active_learning}.
Crucially, it excludes the self-labelling process (stage 2b), in stage 2, thus relying solely on labels from the annotators.
This variant serves as a baseline to evaluate the efficacy of our confidence estimation model in an active learning context, without the influence of self-supervision.
For brevity, we report the CAML results for the translation experiments only (similar trends were observed in the dialogue belief tracking task).

\paragraph{CAMEL} \textbf{C}onfidence-based \textbf{A}cquisition \textbf{M}odel for \textbf{E}fficient self-supervised active \textbf{L}earning is the complete approach, which also includes the self-supervision component.
This variant assesses the value added by self-supervision to the framework, while retaining stages 1, 2, and 4.

\paragraph{CAMELL} \textbf{C}onfidence-based \textbf{A}cquisition \textbf{M}odel for \textbf{E}fficient self-supervised active \textbf{L}earning with \textbf{L}abel validation is an extended variation of our approach that includes a label validation component, denoted as Stage 3 in Section~\ref{subsection:camell:active_learning}.

% --------------------------------------------------------------------------------------
\subsection{Variants of label correction}
\label{subsection:experiments:label_correction}
% --------------------------------------------------------------------------------------

\paragraph{Live label correction} involves simultaneous labelling, validation, and correction of data.
A variant of CAMELL is employed, in which the label is corrected at the validation stage using the prediction of the learning model.

\paragraph{On-line label correction} is a method that labels and validates data simultaneously, with the objective of minimising human effort in providing labels while concurrently validating them.
CAMELL can be employed to flag the data points requiring correction, as well as to apply corrections to the flagged labels using the final model after active learning has been performed.

\paragraph{Offline label correction} is a technique used to correct an already labelled corpus, with the objective of identifying potentially incorrect labels and providing alternatives.
To achieve this, individually trained components of CAMELL can be utilised, specifically the prediction confidence model (Section~\ref{subsubsection:camell:confidence_estimation:prediction}) and the label confidence model (Section~\ref{subsubsection:camell:confidence_estimation:label}). The process consists of the following steps:

\begin{enumerate}
    \item Train learning model on labelled corpus.
    
    \item Generate noisy dataset using this model, leveraging ensemble members from Step 1.If computational constraints prevent the use of an ensemble, a noisy dataset can be generated from a single model using the strategy described in Section~\ref{subsection:camell:efficient_confidence_estimation}.
    
    \item Train learning model on noisy dataset.
    
    \item Train prediction and label confidence models.
    
    \item Perform label correction.
\end{enumerate}

\paragraph{Semi-offline label correction} is a method in which data is collected with the objective of minimising human effort in providing labels, with validation occurring subsequently.
For this purpose, CAMEL can be utilised alongside a separately trained label confidence model (Steps 2 and 3 from above), followed by Step 5.

% --------------------------------------------------------------------------------------
\subsection{Generative Language Modelling Task}
\label{subsection:experiments:genlm}
% --------------------------------------------------------------------------------------

For the generative language modelling task, we explore the application of our CAMEL framework to the task of Neural Machine Translation (NMT).
NMT focuses on converting sequences of text from a source language to a target language.
Our approach involves iterative annotation methods similar to those used in automatic speech recognition~\citep{sperber2016iterativeasr}, which incrementally increase model precision.

Specifically, in our experiment, an annotator corrects individual words within a translation, thereby progressively enhancing the quality of the subsequent output generated by the model.
Conventional annotation methods typically involve providing fully corrected translations or quality ratings.
While this iterative process diverges from conventional methods of machine translation annotation, it allows us to effectively demonstrate the self-supervision mechanism within our framework.

% Note that we refer to this as a feasibility study not to undermine its validity, but to emphasise the exploratory nature of applying our CAMEL framework in an idealised setting.
% This setting provides a clear and controlled environment to evaluate the potential and robustness of our approach before extending it to a more complex scenario.
% Both the NMT and following dialogue belief tracking experiments (Section~\ref{subsection:experiments:dst_task}) are equally important for validating the efficacy of our methodology across different tasks.

% --------------------------------------------------------------------------------------
\subsubsection{Implementation Details}
\label{subsubsection:experiments:genlm:implementation}
% --------------------------------------------------------------------------------------

\begin{figure*}[t]
    \centering
    \includegraphics[scale=0.57]{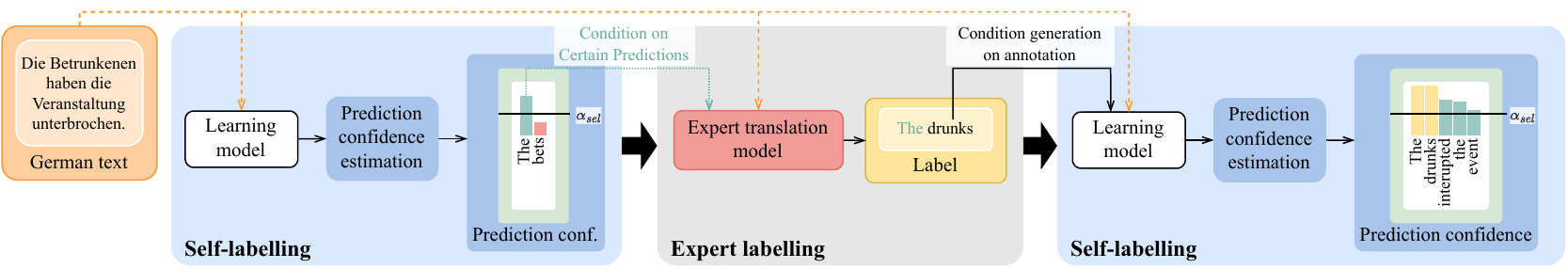}

    \caption{\small The model-based annotation process for semi-supervised annotation for NMT.
    The learning model initiates the translation with the word ``The'', then confidence for the next token generation is below the threshold.
    The expert annotation model is prompted and provides the next word, ``drunks''.
    The learning model resumes and successfully generates the remainder of the translation: ``interrupted the event''.}
    \label{figure:experiments:translation_annotation}
\end{figure*}

We apply CAMEL to the task of machine translation, specifically using the T5 encoder-decoder transformer model (\texttt{t5-small})~\citep{raffel2020t5model}.
We utilise an ensemble of $10$ models in order to produce a well-calibrated predictive distribution, which requires $2500$ GPU hours to fully train.
Approximately $40\%$ of this time is for training the ensemble, $50\%$ for the annotation process, and $10\%$ for training the confidence estimator.

The ensemble model produces two types of uncertainty within the translation process.
The first, termed total uncertainty, is measured by the entropy across the ensemble's predictive distribution.
The second, knowledge uncertainty, is measured by the mutual information shared between the predictive distribution and the individual ensemble models.
These uncertainties are crucial for evaluating the reliability of translations.
The mathematical formulations for calculating total $(\mathcal{T})$ and knowledge $(\mathcal{K})$ uncertainties are as follows:
%
% Note: We use the summing index 'j' consistently to define the average of the ensemble
\begin{empheq}{align*}
    \begin{split}
        \mathcal{T} \left( \boldpi \right) 
        & 
        = \mathcal{H} \left[ \frac{1}{\EnsembleSize} \sum\nolimits_{j=1}^\EnsembleSize \boldpi^{(j)} \right],
        \\
        \mathcal{K} \left( \boldpi \right) & = \frac{1}{\EnsembleSize} \sum\nolimits_{\EnsembleIndex=1}^\EnsembleSize D_\texttt{KL} 
        \left[ 
            \boldpi^{(\EnsembleIndex)} \Big\Vert \frac{1}{\EnsembleSize} \sum\nolimits_{j=1}^\EnsembleSize \bm{\pi}^{(j)} 
        \right],
    \end{split}
\end{empheq}
where $\bm{\pi}^{(\EnsembleIndex)}$ represents the predictive distribution from the $\EnsembleIndex\textsuperscript{th}$ ensemble member.

The WMT17 DE-EN dataset, which consists of German to English translations~\citep{bojar2017wmt17}, is used for training, and METEOR~\citep{banerjee2005meteor}, BLEU~\citep{papineni2002bleu}, and COMET~\citep{rei2020comet} serve as evaluation metrics.

\begin{figure*}[h!]
    \centering

    \begin{subfigure}[b]{0.49\textwidth}
        \includegraphics[scale=0.42]{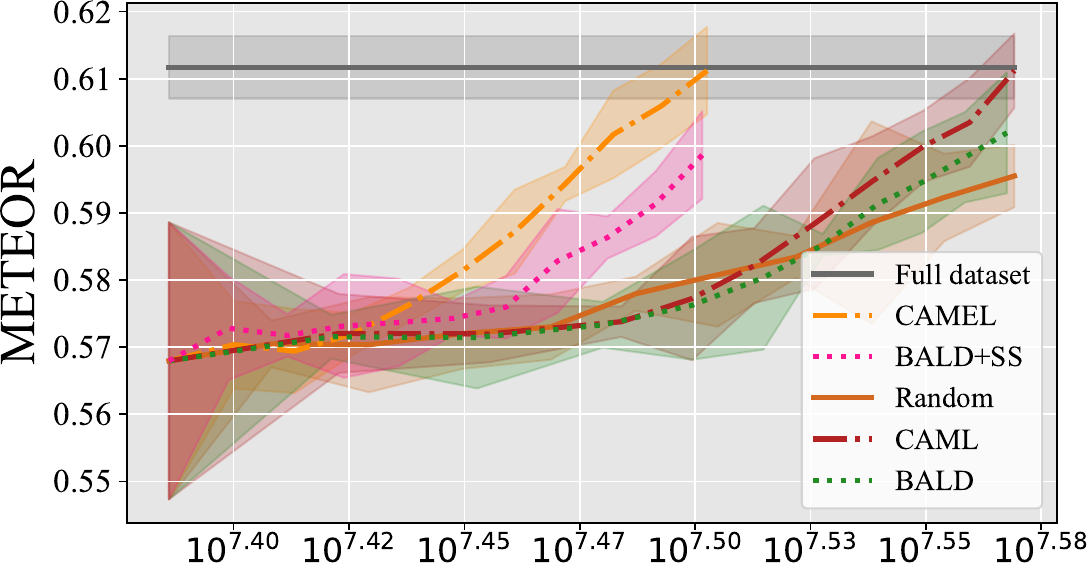}

        \subcaption{\small Number of word-level labels}
        \label{subfigure:experiments:translation_results:n_annotations}
    \end{subfigure}
    \begin{subfigure}[b]{0.49\textwidth}
        \includegraphics[scale=0.42]{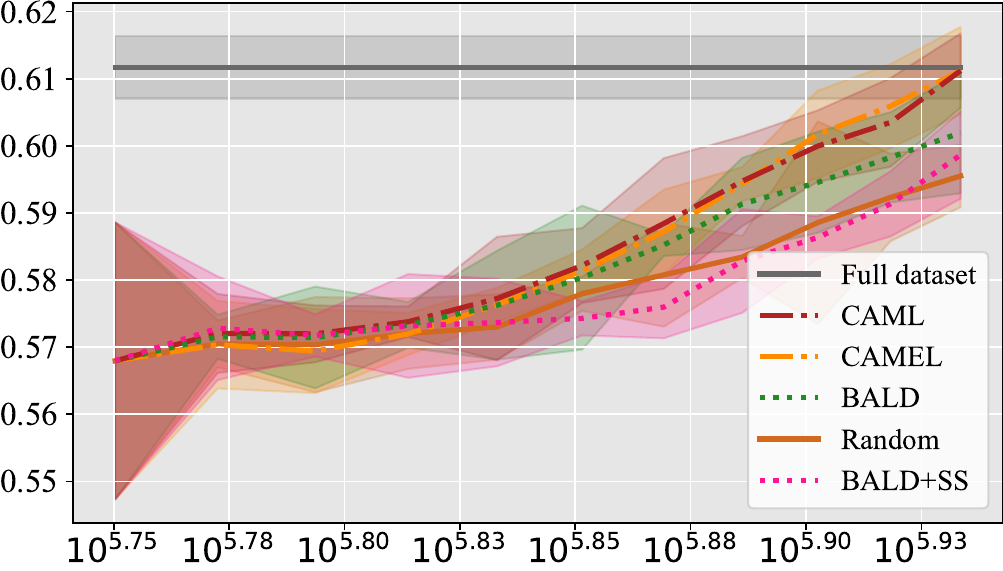}

        \subcaption{\small Number of complete translations}
        \label{subfigure:experiments:translation_results:n_translations}
    \end{subfigure}

    \caption{\small METEOR score of the T5 translation model using different active learning approaches on the WMT$17$ DE-EN test set, as a function of (a) the number of word-level labels and (b) the number of complete translations, with $95\%$ confidence interval.}
    \label{figure:experiments:translation_results}
\end{figure*}

\begin{figure*}[h!]
    \centering

    \begin{subfigure}[b]{0.49\textwidth}
        \includegraphics[scale=0.42]{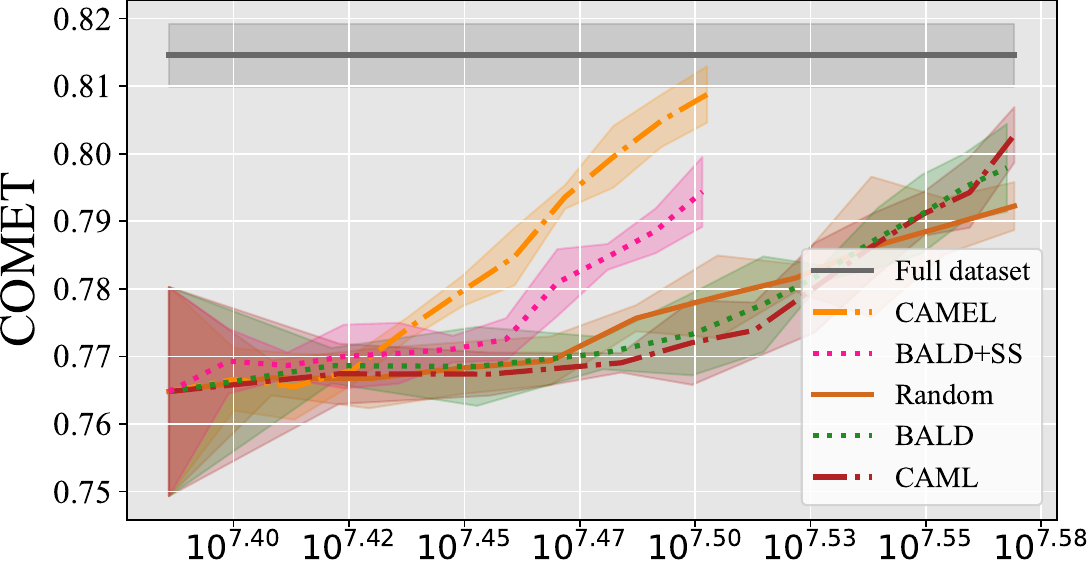}

        \subcaption{\small Number of word-level labels}
        \label{subfigure:experiments:translation_results_comet:n_annotations}
    \end{subfigure}
    \begin{subfigure}[b]{0.49\textwidth}
        \includegraphics[scale=0.42]{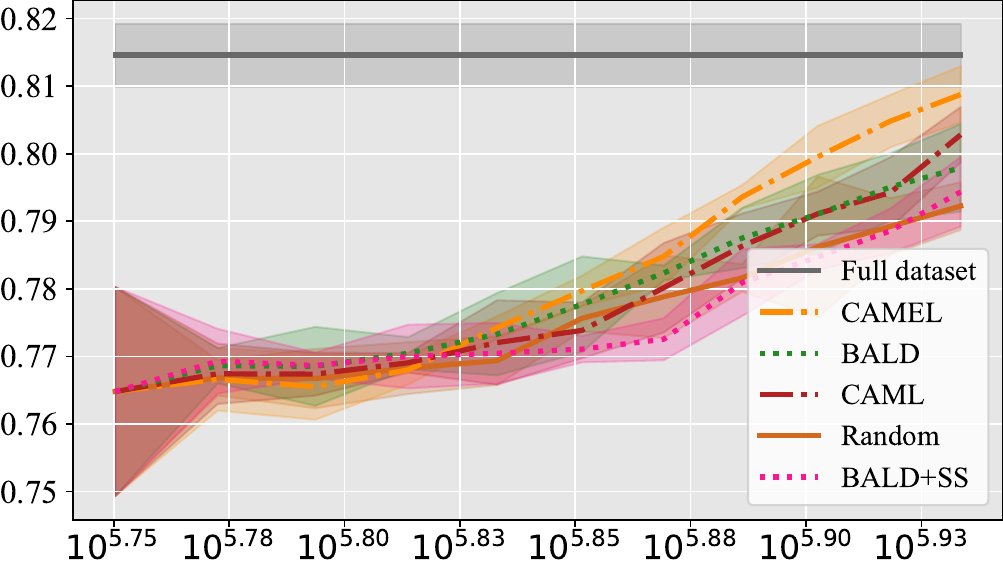}

        \subcaption{\small Number of complete translations}
        \label{subfigure:experiments:translation_results_comet:n_translations}
    \end{subfigure}

    \caption{\small COMET score of the T5 translation model using different active learning approaches on the WMT$17$ DE-EN test set, as a function of (a) the number of word-level labels and (b) the number of complete translations, with $95\%$ confidence interval.}
    \label{figure:experiments:translation_results_comet}
\end{figure*}

As machine translation does not entail a multi-output task, we employed a simplified version of the confidence estimation model, introduced in Section~\ref{subsection:camell:confidence_estimation}, consisting of only the intra-category encoder.
The latent dimension of the encoder and feature transformation layer is $16$.
The parameters are optimised using the standard binary negative log likelihood loss~\citep{cox1958regression}.

It is crucial to address the inherent challenges in sequential machine translation labelling: (1) future sentence structure and labels can change depending on the current label, and (2) for any word position there exist multiple valid candidate words.
This complexity necessitates the use of a dynamic annotation approach, as static dataset labels are insufficient for new data labeling.
To avoid high translation annotation costs, we propose a practical approach: using an \emph{expert} translation model, specifically the MBART-50 multilingual model~\citep{tang2020mbart50}, to simulate a human annotator.

Our approach, depicted in Figure~\ref{figure:experiments:translation_annotation}, is a multi-stage procedure.
Initially, the learning model produces a translation for a selected source language sentence.
As it generates the translation, it simultaneously estimates its confidence for the subsequent token.
Should this confidence fall below a set threshold $\alpha_{\text{sel}}$, the \emph{expert} translation model steps in to supply the next word in the translation.
After the label is provided, the learning model resumes the translation generation.
For any future token whose confidence drops below the threshold, the \emph{expert} translation model re-engages.
This process continues until a complete translation for the source sentence is realised.
The uncertainty threshold, $\alpha_{\text{sel}}$, is strategically chosen to yield a maximum of $N_{\text{ann}}$ word labels.

% Regarding the exclusion of label validation (stage $3$ in section~\ref{subsection:camell:active_learning}) in the MT experiments, the primary reason is the challenge in applying it to NMT.
% The label validation phase requires generating noisy labels, typically done using the predictive distributions of an ensemble of models.
% However, producing human-like noisy text, rather than random or nonsensical text, for machine translation data presents a challenge.
% Developing a realistic noise model for such data is necessary but is considered beyond the scope of the present study.
% We propose this as a direction for future research, emphasizing the exploration of methods for effectively integrating label validation into neural machine translation (NMT).

% Regarding the exclusion of label validation (stage $3$ in section~\ref{subsection:camell:active_learning}) in the MT experiments, the primary reason is the challenge in applying it to NMT.
% The label validation phase requires generating noisy labels, typically done using the predictive distributions of an ensemble of models.
% However, generating human-like noisy text, rather than random meaningless text, for translation data presents a non-trivial challenge.
% Consequently, we consider this to be beyond the scope of the current study and propose it as future work to explore methods for effectively integrating label validation in neural machine translation (NMT).

% --------------------------------------------------------------------------------------
\subsubsection{Results}
\label{subsubsection:experiments:genlm:results}
% --------------------------------------------------------------------------------------

We evaluated the performance of our proposed CAMEL framework and baseline models using METEOR~\citep{banerjee2005meteor}, BLEU~\citep{papineni2002bleu}, and COMET~\citep{rei2020comet} scores.
While traditional metrics such as METEOR and BLEU highlight similar trends (with BLEU scores included in Appendix~\ref{appendix:bleu_scores}), COMET, a neural evaluation metric, provides a more comprehensive understanding of the translation quality beyond traditional metrics.
% }
We establish that our proposed CAMEL framework, enhanced with self-supervision, is significantly more efficient requesting word-level labels than baseline models like BALD, BALD+SS, and random selection.
This efficiency is evident in Figures~\ref{subfigure:experiments:translation_results:n_annotations} and~\ref{subfigure:experiments:translation_results_comet:n_annotations}, which showcases CAMEL's need for fewer word-level labels to achieve similar performance.
Although our primary focus is on the number of word-level labels queried, it is crucial to note that labelling overhead is also accounted for. We measure this overhead by the effort required to read and understand the source language tokens, which we consider a sufficient indicator.

A notable point to observe in Figure~\ref{subfigure:experiments:translation_results:n_translations} is that the introduction of self-supervision to CAMEL does not significantly influence its performance in terms of the number of complete translations required, as evident by the comparison between CAML (CAMEL without the self-supervised labelling component) and CAMEL.
This implies that self-supervision within CAMEL is applied predominantly when the model's predictions can be considered reliable.
In contrast, we observe that BALD+SS, despite its label efficiency shown in Figure~\ref{subfigure:experiments:translation_results:n_annotations}, performs poorly in terms of the number of complete translations required, as demonstrated in Figure~\ref{subfigure:experiments:translation_results:n_translations}.
This drop in performance may be attributed to BALD+SS's tendency to incorrectly self-label complex examples.
This trend is further evidenced by CAML's lower expected calibration error (ECE), reported in Table~\ref{table:experiments:ece}.
The COMET results, presented in Figure~\ref{figure:experiments:translation_results_comet}, further attest to CAMEL's superiority.
CAMEL not only excels in reducing the number of word-level labels but also outperforms other models in the number of complete translations required.
The non-overlapping confidence intervals in the results indicates that the improvements of CAMEL over other methods are statistically significant.

Regardless of the methodology used, all models require roughly the same number of complete translations, as shown in Figures~\ref{subfigure:experiments:translation_results:n_translations} and~\ref{subfigure:experiments:translation_results_comet:n_translations}.
This supports the widely accepted notion that exposure to large datasets is vital for training robust natural language processing (NLP) models.

Encouraged by these results, we adapt CAMEL to address the dialogue belief tracking problem, a  task plagued by errors in the labels of available datasets.
% --------------------------------------------------------------------------------------
\subsection{Dialogue Belief Tracking Task}
\label{subsection:experiments:dst_task}
% --------------------------------------------------------------------------------------

In task-oriented dialogue, the dialogue ontology $\mathcal{O}$ contains a set of $M$ \texttt{domain-slot} pairs $\{s^1,s^2,\ldots,s^M\}$ and a set of plausible values $\mathcal{V}_{s^m}$ for each $s^m$.
The goal of the dialogue belief tracker is to infer the user's preference for each $s^m$ by predicting a probability distribution over the plausible values.
Notably, each set of plausible values, $\mathcal{V}_{s^m}$, includes the \texttt{not\_mentioned} value, indicating that a specific \texttt{domain-slot} pair is not part of the user's goal~\citep{feng-etal-2024-infusing,geishauser2024openhorizen}.
This allows for computing the model's confidence for slots not present in the user's preference.

To train a belief tracking model, we require the dialogue state, which includes the \texttt{value} label for each \texttt{domain-slot}, in every dialogue turn.
The dialogue state at turn $t$ in dialogue $i$ is represented as $\mathcal{B}_{i, t} = \{ (s^m, v_{i, t}^{s^m}) \}_{s^m \in \mathcal{O}}$, where $v_{i, t}^{s^m}$ denotes the \texttt{value} for the \texttt{domain-slot} pair $s^m$ at turn $t$ in dialogue $i$.
Consequently, we obtain a dataset 
$\mathcal{D} = \{
    (\mathbf{u}_{i, 1:t}^{\text{usr}}, 
    \mathbf{u}_{i, 1:t-1}^{\text{sys}}, 
    \mathcal{B}_{i, t})_{t=1}^{T_i}
\}_{i=1}^N$, 
consisting of $N$ dialogues, each comprising $T_i$ turns, where user and system utterances at turn $t$ in dialogue $i$ are denoted as $\mathbf{u}_{i, t}^{\text{usr}}$ and $\mathbf{u}_{i, t}^{\text{sys}}$, respectively.

To create a dataset $\mathcal{D}$, annotators usually provide relevant values for the \texttt{domain-slot} pairs they believe are present in the user's utterance at every turn $t$.
Subsequently, a handcrafted rule-based tracker considers the previous state $\mathcal{B}_{i, t-1}$, the semantic actions present in the system utterance and the values provided by the annotator to generate complete dialogue states for each turn~\citep{budzianowski2018multiwoz}.
However, this approach has several drawbacks.
Firstly, rule-based trackers tend to be imprecise and necessitate redevelopment for each new application, making it less versatile~\citep{vukovic2024ontology}.
Secondly, it may not use the time of human annotators efficiently, as the learning model could potentially predict the state for a substantial part of the dialogue accurately.
Lastly, there is the risk of human annotators inadvertently overlooking slots in the user input, which could result in incomplete data.
\begin{table}[t]
    \centering
    \resizebox{0.8\columnwidth}{!}{%
    \begin{tabular}{ccc}
        \toprule
        \textbf{Confidence Estimator} & \textbf{Dataset} & \textbf{ECE (\%)  $\downarrow$} 
        \\
        \midrule
        CE-T5 + CAML & WMT$17$ DE-EN & $\bm{26.74*}$ \\
        CE-T5 + BALD & WMT$17$ DE-EN & $47.21$ \\
        \midrule
        CE-SetSUMBT + CAML & MultiWOZ $2.1$ & $\bm{9.65*}$ \\
        CE-SetSUMBT + BALD & MultiWOZ $2.1$ & $17.21$ \\
        \bottomrule
    \end{tabular}%
    }

    \caption{\small Comparison of the expected calibration error (ECE) of confidence estimation approaches.
    $\bm{*}$ indicates significant difference on $95\%$ confidence interval.}
    \label{table:experiments:ece}
\end{table}

% --------------------------------------------------------------------------------------
\subsubsection{Learning Model}
\label{subsubsection:experiments:dst_task:learning_model}
% --------------------------------------------------------------------------------------

To apply \emph{CAMEL} to the dialogue belief tracking problem, we use the CE-SetSUMBT (Calibrated Ensemble -- SetSUMBT) model~\citep{vanniekerk2021setsumbt}, a model which produces well-calibrated uncertainty estimates, important for CAMEL.
The CE-SetSUMBT model consists of $10$ ensemble members, requiring $1000$ GPU hours to fully train.
Approximately $45\%$ of this time is utilised for training the ensemble, $45\%$ for training the \emph{noisy} model, and $10\%$ for training the confidence estimators.
In addition, we integrate the post-hoc uncertainty learning using a Dirichlet meta-model approach~\citep{shen2023metauq}, described in Section~\ref{subsection:camell:efficient_confidence_estimation}, into SetSUMBT.

% --------------------------------------------------------------------------------------
\subsubsection{Datasets}
\label{subsubsection:experiments:dst_task:datasets}
% --------------------------------------------------------------------------------------

In order to test our proposed approach, we utilise the multi-domain task-oriented dialogue dataset MultiWOZ 2.1~\citep{eric2019multiwoz21,budzianowski2018multiwoz} and its manually corrected test set provided in MultiWOZ 2.4~\citep{ye2022multiwoz24}.
In our experiments, we regard MultiWOZ 2.1 as a dataset with substantial label noise~\citep{eric2019multiwoz21,zang2020multiwoz22, ye2022multiwoz24}, and the test set of MultiWOZ 2.4 a dataset with accurate labels.

% --------------------------------------------------------------------------------------
\subsubsection{Implementation Details}
\label{subsubsection:experiments:dst_task:implementation}
% --------------------------------------------------------------------------------------

\begin{figure*}[t!]
    \centering

    \begin{subfigure}[b]{0.49\textwidth}
        \includegraphics[scale=0.42]{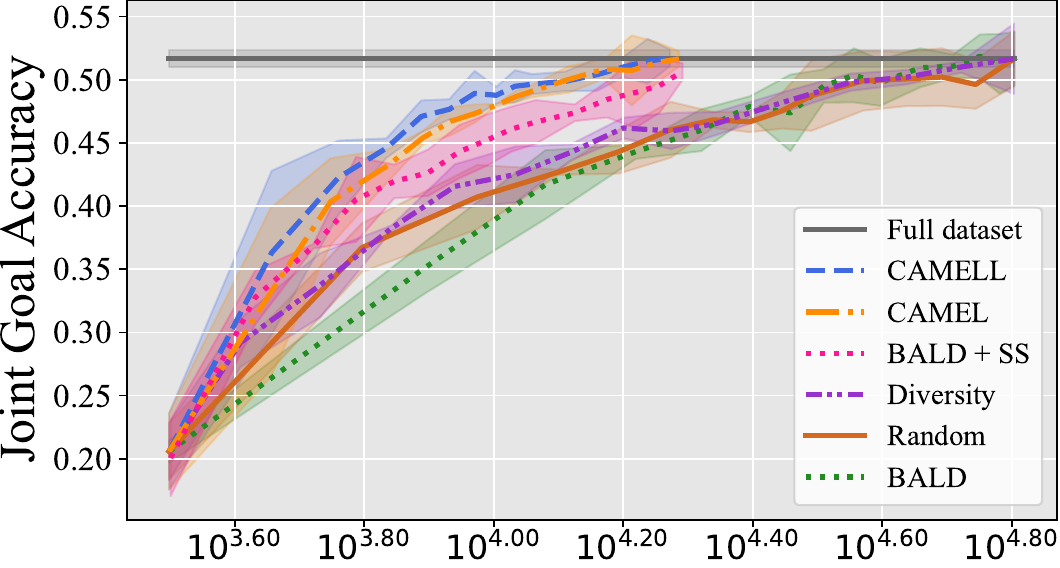}

        \subcaption{\small Number of labels}
        \label{subfigure:experiments:dst_results:n_annotations}
    \end{subfigure}
    \begin{subfigure}[b]{0.49\textwidth}
        \includegraphics[scale=0.42]{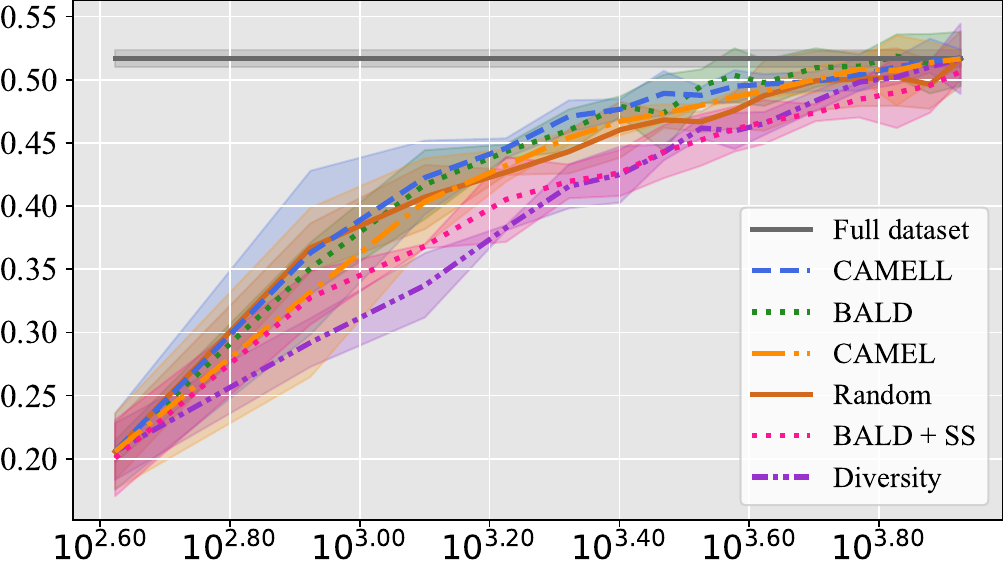}

        \subcaption{\small Number of dialogues}
        \label{subfigure:experiments:dst_results:n_dialogues}
    \end{subfigure}

    \caption{\small JGA of the CE-SetSUMBT model using different active learning approaches, on the MultiWOZ $2.1$ test set, as a function of (a) the number of labels and (b) the number of dialogues, with $95\%$ conf. int.}
    \label{figure:experiments:dst_results}
\end{figure*}

\begin{figure*}[t!]
    \centering
    \begin{subfigure}[b]{0.49\textwidth}
        \includegraphics[scale=0.42]{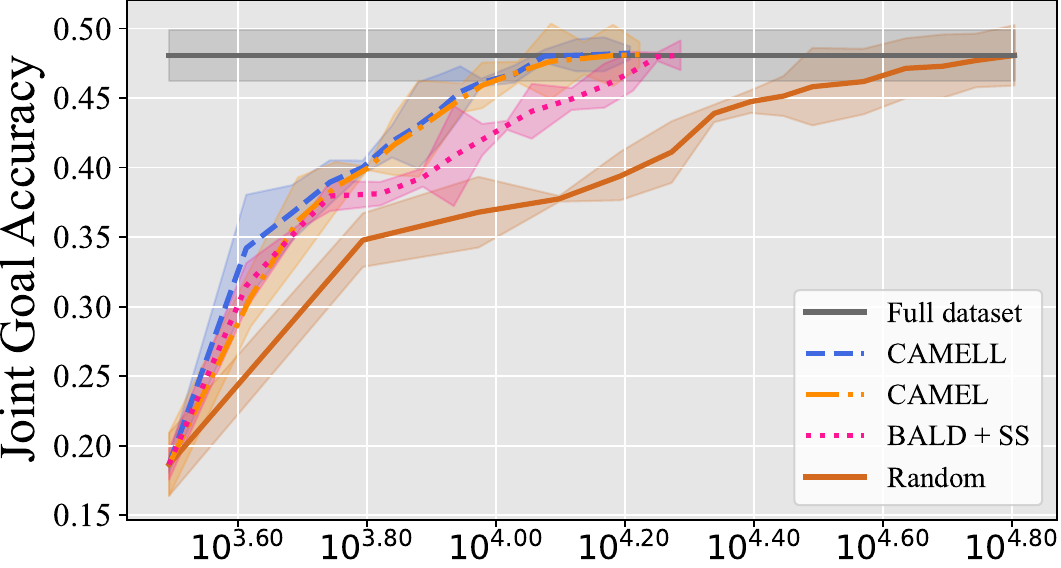}
        \subcaption{\small Number of labels}
        \label{subfigure:experiments:dst_results:meta_n_annotations}
    \end{subfigure}
    \begin{subfigure}[b]{0.49\textwidth}
        \includegraphics[scale=0.42]{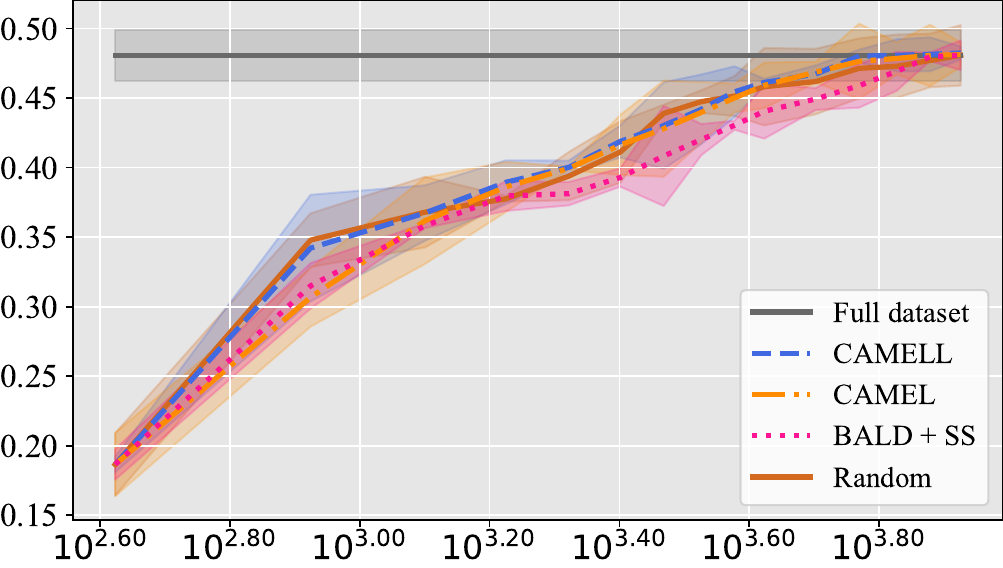}
        \subcaption{\small Number of dialogues}
        \label{subfigure:experiments:dst_results:meta_n_dialogues}
    \end{subfigure}
    \caption{\small JGA of the Dirichlet Meta SetSUMBT model using different active learning approaches, on the MultiWOZ $2.1$ test set, as a function of (a) the number of labels and (b) the number of dialogues, with $95\%$ conf. int.}
    \label{figure:experiments:meta_dst_results}
\end{figure*}

The latent dimension of the intra- and inter-category encoders and feature transformation layer is $16$.
During training of the label confidence estimation model (Section~\ref{subsubsection:camell:confidence_estimation:label}), to avoid overfitting, we improve the calibration of this model by deploying binary label smoothing loss~\citep{szegedy2016ls}, temperature scaling and noisy training using Gaussian noise~\citep{an1996noisytraining}.

For the \emph{seed} dataset (Section~\ref{section:camell}) we randomly select $5\%$ of dialogues on which we train the initial SetSUMBT model.
The other dialogues in the dataset are treated as the unlabelled pool.
At each update step another $5\%$ of the data are selected to be labelled.
At each point where we require expert labels, we take the original labels provided in the dataset to simulate a human annotator.

% --------------------------------------------------------------------------------------
\subsubsection{Evaluation}
\label{subsubsection:experiments:dst_task:evaluation}
% --------------------------------------------------------------------------------------

As the main metric for our experiments, we use joint goal accuracy (JGA)~\citep{henderson2014dstc}.
We further include the joint goal expected calibration error (ECE)~\citep{guo2017temp, vanniekerk2020cebst}, which measures the calibration of the model.
In terms of measuring efficiency of each method, we examine JGA as a function of the number of expert provided labels.
In order to assess the quality of the corrected dataset, we measure the JGA of models trained on a noisy dataset, with and without the proposed label correction.

% --------------------------------------------------------------------------------------
\subsubsection{Dialogue Diversity Baseline}
\label{subsubsection:experiments:dst_task:diversity_baseline}
% --------------------------------------------------------------------------------------

We include an additional dialogue diversity baseline, aiming to obtain labels for dialogues geometrically dissimilar from those in the labelled pool, thus ensuring data space coverage.
This diversity strategy proposed by~\citet{xie2018cost} assesses similarity based on vector embeddings of the candidate dialogue versus labelled dialogues.
We adapt this approach by employing RoBERTa model embeddings~\citep{liu2019roberta}, fine-tuned in an unsupervised fashion, on the MultiWOZ dialogues.

% --------------------------------------------------------------------------------------
\subsubsection{Results}
\label{subsubsection:experiments:dst_task:results}
% --------------------------------------------------------------------------------------

As shown in Figure~\ref{subfigure:experiments:dst_results:n_annotations}, our proposed CAMEL framework requires significantly fewer labels to reach performance levels comparable to those of the baseline methods.
This indicates that CAMEL is more efficient in learning dialogue belief tracking than the baseline strategies.
It is important to note that all approaches requires the same number of unlabelled dialogues (see Figure~\ref{subfigure:experiments:dst_results:n_dialogues}).
It also highlights the role played by CAMEL's confidence estimates in guiding the active learning process.
This conclusion is supported by the lower calibration error of CAMEL's confidence estimates, as reported in Table~\ref{table:experiments:ece}.

Further, we observe in Figures~\ref{subfigure:experiments:dst_results:meta_n_annotations}-\ref{subfigure:experiments:dst_results:meta_n_dialogues} that similar results can be achieved using a computationally efficient uncertainty estimation technique such as the post-hoc Dirichlet meta model, described in Section~\ref{subsection:camell:efficient_confidence_estimation}, applied to the SetSUMBT model.
It should be noted that the comparatively lower joint goal accuracy of this model can be attributed to its singular SetSUMBT model configuration.
An ensemble of models consistently achieves an accuracy that is \(2\) to \(3\) percentage points higher.

% --------------------------------------------------------------------------------------
\subsection{Label Correction}
\label{subsection:experiments:label_correction}
% --------------------------------------------------------------------------------------

To assess the quality of the corrected labels generated by our proposed label correction method (Section~\ref{subsection:camell:label_correction}), we trained two distinct tracking models, CE-SetSUMBT and TripPy~\citep{heck2020trippy}, using both the original MultiWOZ $2.1$ dataset and various autocorrected datasets (live, online, offline, and semi-offline).
The evaluation was conducted on both the noisy MultiWOZ $2.1$ test set and the manually corrected MultiWOZ $2.4$ test set.
The selected tracking models represent the two major non-generative approaches to dialogue state tracking: a pick-list-based approach (SetSUMBT) and a span-prediction approach (TripPy).

% --------------------------------------------------------------------------------------
\subsubsection{Results}
\label{subsubsection:experiments:label_correction:results}
% --------------------------------------------------------------------------------------

\begin{table}[t]
    \centering
    \resizebox{\columnwidth}{!}{%
    \begin{tabular}{llcc}
        \toprule
        \textbf{Model} &
        \begin{tabular}{@{}l@{}}\textbf{Label Corr.}\\\textbf{Setup}\end{tabular} &
        \textbf{MultiWOZ 2.1} &
        \textbf{MultiWOZ 2.4} \\
        \midrule
        \multirow{5}{*}{CE-SetSUMBT}    & None          & $51.79$   & $61.63$ \\
                                        & Live          & $32.48$   & $37.32$ \\
                                        & Online        & $52.85$   & $\bm{63.35}$ \\
                                        & Offline       & $52.83$   & $\bm{63.32*}$ \\
                                        & Semi-offline  & $52.69$   & $63.12$ \\
        \midrule
        \multirow{4}{*}{TripPy}         & None          & $55.28$   & $64.45$ \\
                                        & Online        & $56.17$   & $\bm{66.13}$ \\
                                        & Offline       & $56.11$   & $\bm{66.02*}$ \\ 
                                        & Semi-offline  & $55.85$   & $65.82$ \\
        \bottomrule
    \end{tabular}%
    }
    \caption{\small Comparison of JGA of trackers trained with and without label corrections.
    The label corrections can be obtained using a SetSUMBT model trained on the full MultiWOZ $2.1$ dataset, trained using CAMEL, or trained using CAMELL.
    $\bm{*}$ indicates significant difference on $95\%$ conf. int.
    \label{table:experiments:label_correction}}
\end{table}

In Table~\ref{table:experiments:label_correction}, we present the JGA of the CE-SetSUMBT models on two test sets: the (noisy) MultiWOZ $2.1$ test set and the (manually corrected) MultiWOZ $2.4$ test set.\footnote{The MultiWOZ $2.4$ validation set was never used during training.}
Overall, results show the same trend both for CE-SetSUMBT and TripPy: on the MultiWOZ $2.1$ test set, the models do not show statistically significant improvements, which is unsurprising given that the MultiWOZ $2.1$ test set contains errors and, therefore, cannot adequately assess the impact of label correction.
In contrast, on the MultiWOZ $2.4$ test set, we observe significant improvements for both offline and online label correction methods for both belief state trackers.
This demonstrates that the datasets resulting from online and offline label correction are of significantly higher quality.

The semi-offline method fails to produce significant improvements.
We hypothesise that the model trained using CAMEL has already acquired similar error patterns to those commonly made by human annotators.
The live label correction setup results in a low-quality dataset, which we attribute to the model's inherent inability to correct data selected through active learning.
At this stage, the model lacks the capability to make accurate predictions for these instances.\footnote{This method is not examined for TripPy, as we do not expect it to behave differently.}

\begin{table}[t]
    \centering
    \resizebox{\columnwidth}{!}{%
    \begin{tabular}{clc}
        \toprule
        \begin{tabular}{@{}l@{}}Error\\Type\end{tabular} & Conversation & \begin{tabular}{@{}c@{}}MultiWOZ $2.1$ Labels and\\ Corrections\end{tabular}
        \\
        \midrule
        I & \begin{tabular}{@{}l@{}}\emph{User:} I would like to find a\\ place that serves \textbf{moderately}\\ priced \textbf{Chinese} food.\end{tabular} & \begin{tabular}{@{}l@{}}\texttt{\{Restaurant:}\\\hspace{0.4cm}\texttt{\{Food: Chinese,} $(95\%)$\\\hspace{0.6cm}\texttt{Price: Moderate,} $(94\%)$\\\color{red}\hspace{0.6cm}\texttt{Day: Tuesday,} $(11\%)$\\\color{DarkBlue}\hspace{0.6cm}\texttt{Day: not\_mentioned\}\}} $(72\%)$\end{tabular} 
        \\
        \midrule
        II & \begin{tabular}{@{}l@{}}\emph{User:} I feel like going to a\\ \textbf{nightclub}.\\\emph{System:} Okay, the \textbf{Soul Tree}\\ \textbf{Nightclub} is a popular place.\\ Would you like the address\\or phone number?\\\emph{User:} I will appreciate that.\end{tabular} & \begin{tabular}{@{}l@{}}\texttt{\{Attraction:}\\\hspace{0.4cm}\texttt{\{Type: Night club,} $(94\%)$\\\hspace{0.6cm}\texttt{Name: Soul Tree\},} $(53\%)$\\\hspace{0.2cm}\texttt{Hotel:}\\\color{red}\hspace{0.4cm}\texttt{\{Name: Sou,} $(14\%)$\\\color{DarkBlue}\hspace{0.6cm}\texttt{Name: not\_mentioned\}\}} $(34\%)$\end{tabular} \\
        \midrule
        III & \begin{tabular}{@{}l@{}}\emph{User:} I need a train leaving\\ on \textbf{Friday} and I want to get\\there by $\bm{21:30}$. Leaving\\\textbf{Broxbourne}.\end{tabular} & \begin{tabular}{@{}l@{}}\texttt{\{Train:}\\\hspace{0.4cm}\texttt{\{Dept.: Broxbourne,} $(94\%)$\\\hspace{0.6cm}\texttt{Day: Friday,} $(95\%)$\\\color{red}\hspace{0.6cm}\texttt{Arrive by: 21:20,} $(1\%)$\\\color{DarkBlue}\hspace{0.6cm}\texttt{Arrive by: 21:30\}\}} $(83\%)$\end{tabular}
        \\
        \bottomrule
    \end{tabular}%
    }
    \caption{\small Examples of three common types of annotation errors in the MultiWOZ $2.1$ dataset detected and corrected by CAMELL, (I) hallucinated annotations, (II) multi-annotation and (III) erroneous annotation.
    For each, we provide the confidence scores of the labels and the corrections proposed by the model.
    Incorrect labels are marked in red and the proposed corrections in blue.}
    \label{figure:experiments:label_correction_examples}
\end{table}

Although the label validation stage of CAMELL does not yield a statistically significant improvement in the active learning setting, it produces a model that provides more reliable label correction compared to the CAMEL approach without label validation (see online vs. semi-offline correction in Table~\ref{table:experiments:label_correction}).
While CAMELL does not generate labels of higher quality than those produced by the offline label correction approach, it facilitates the creation of a clean dataset with fewer labels, thereby reducing human effort.

An important take-away message is: if all labels in the dataset are available and active learning is not required, offline label correction can be applied to enhance the dataset's quality.
However, if labels are being collected through an active learning process, an online label correction should be applied rather than a semi-offline method, as the label validation component enables the creation of a final dataset of higher quality.

% --------------------------------------------------------------------------------------
\subsubsection{Qualitative Analysis}
\label{subsubsection:experiments:label_correction:analysis}
% --------------------------------------------------------------------------------------

In our investigation of the improved datasets obtained from offline label correction, we identified three prevalent label errors, which our approach successfully rectifies, as exemplified in Table~\ref{figure:experiments:label_correction_examples}.
(I)~Hallucinated annotations, where the annotator assigns labels not present in the dialogue context, (II)~Multi-annotation, the case of assigning multiple labels to the same piece of information, and (III)~Erroneous annotation, the situation where an incorrect label is assigned based on the context.
These instances underscore the efficacy of our label validation model in minimising the propagation of errors into the dataset.
  % --------------------------------------------------------------------------------------
\section{Conclusion}
\label{section:conclusion}
% --------------------------------------------------------------------------------------

We propose CAMEL, a novel active learning approach that integrates self-supervision, with the goal of minimizing the reliance on labelled data in addressing sequential multi-output labelling problems.
Initially, we applied CAMEL to a generative language modelling task in an idealized setting, specifically focusing on machine translation.
Subsequently, in a more realistic setting focused on the dialogue belief tracking task, we demonstrated that our approach significantly outperforms baseline methods in terms of robustness and data efficiency.

Additionally, we introduce a methodology for automated dataset correction.
Our experiments confirm that our label correction method enhances the overall quality of a dataset.
We demonstrate that CAMELL (with label validation) is capable of producing high-quality datasets with a fraction of the human annotation required, through online label correction, thereby highlighting the importance of the label validation component for this task.

Finally, it is important to note that while many presented experiments used ensembles to establish comparisons, we have also provided a mechanism for confidence estimation and active learning that \emph{does not} utilise ensembles and thus is more environmentally friendly.

We believe that this work has far-reaching implications.
Firstly, it underscores the indispensable role of uncertainty estimation in learning models.
Secondly, the versatility of CAMEL opens up possibilities for its application across diverse sequential multi-output labelling problems, such as entity-relation extraction or weather forecasting.
Thirdly, it demonstrates that, in principle, dataset deficiencies can be addressed via data-driven approaches, circumventing the need for extensive manual or rule-based curation.
This is particularly pertinent considering the prevailing belief that undesirable outcomes produced by NLP models are inherently linked to the training datasets and cannot be rectified algorithmically~\cite[14.6.3]{eisenstein2019introduction}.

Looking ahead, we anticipate that refining the process of generating \emph{noisy} datasets could result in a model capable of not only identifying label noise but also filtering out biases, false premises, and misinformation.
  % --------------------------------------------------------------------------------------
\section{Acknowledgements}
\label{section:acknowledgements}
% --------------------------------------------------------------------------------------

This work was made possible through the support of the Alexander von Humboldt Foundation, provided within the Sofja Kovalevskaja Award, the European Research Council (ERC) under the Horizon 2020 research and innovation program (Grant No. STG2018 804636), and the Ministry of Culture and Science of North Rhine-Westphalia within the Lamarr Fellow Network.
Computational resources were provided by the Centre for Information and Media Technology at Heinrich Heine University Düsseldorf and Google Cloud.
We thank the anonymous reviewers for their insightful comments and suggestions, particularly for encouraging us to develop a more computationally efficient approach to uncertainty estimation.
We also thank Andrey Malinin for early discussions that inspired us to broaden our perspective beyond dialogue state tracking, as well as Prof. Joseph van Genabith for his valuable insights regarding the machine translation setting.

  \bibliography{refs}
  \bibliographystyle{acl_natbib}

  \clearpage
  \newpage
  \appendix
% --------------------------------------------------------------------------------------
\section{BLEU Scores for Translation Experiments}
\label{appendix:bleu_scores}
% --------------------------------------------------------------------------------------

\begin{figure}[h!]
    \centering

    \begin{subfigure}[b]{0.49\textwidth}
        \includegraphics[scale=0.42]{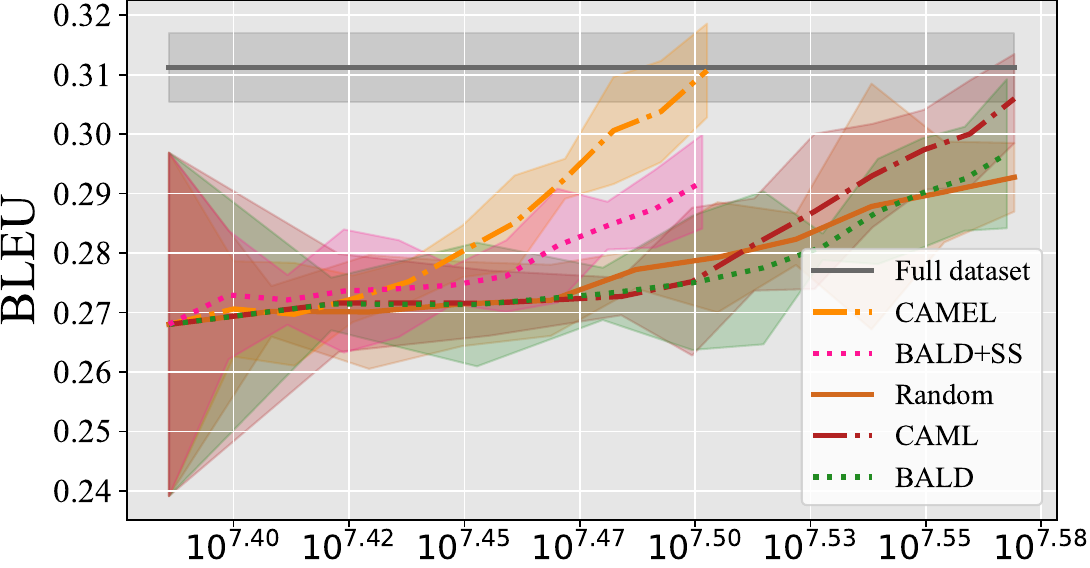}

        \subcaption{\small Number of word-level labels}
        \label{subfigure:experiments:translation_results_bleu:n_annotations}
    \end{subfigure}
    
    \begin{subfigure}[b]{0.49\textwidth}
        \includegraphics[scale=0.42]{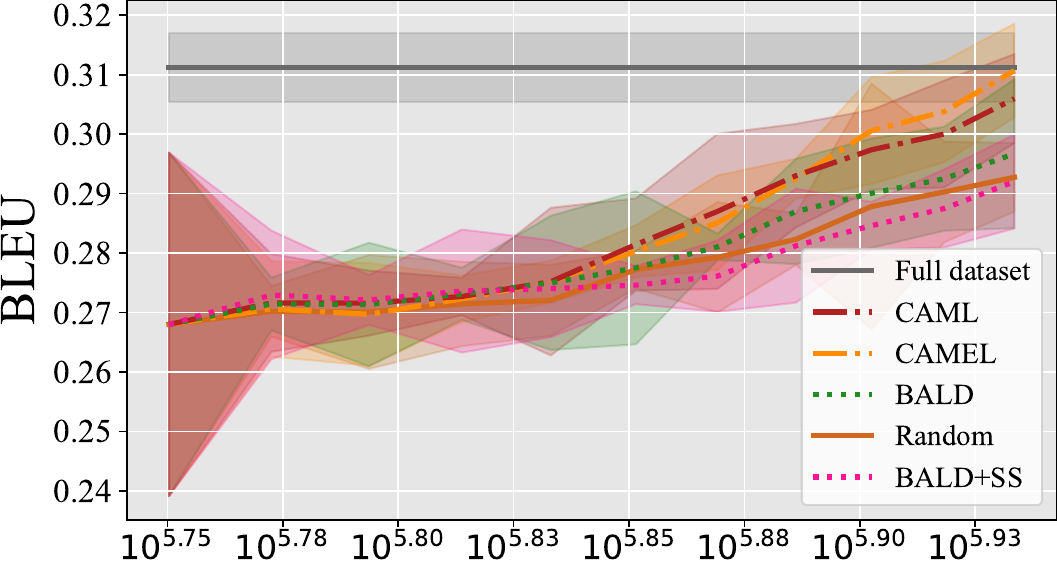}

        \subcaption{\small Number of complete translations}
        \label{subfigure:experiments:translation_results_bleu:n_translations}
    \end{subfigure}

    \caption{\small BLEU score of the T5 translation model using different active learning approaches on the WMT$17$ DE-EN test set, as a function of (a) the number of word-level labels and (b) the number of complete translations, with $95\%$ conf. int.}
    \label{figure:experiments:translation_results_bleu}
\end{figure}
\end{document}